\documentclass[10pt,twocolumn,letterpaper]{article}

\usepackage[pagenumbers]{iccv} 

\definecolor{iccvblue}{rgb}{0.21,0.49,0.74}
\usepackage[pagebackref,breaklinks,colorlinks,allcolors=iccvblue]{hyperref}

\usepackage{makecell}
\usepackage{colortbl}
\usepackage{xcolor}
\usepackage[table]{xcolor}
\usepackage{tabularx}      
\usepackage{booktabs}
\usepackage{multirow}
\usepackage{array}
\usepackage{enumitem}
\usepackage{listings}

\newcommand{\systemname}{{GraspCoT}\xspace}
\newcommand{\datasetname}{{IntentGrasp}\xspace}

\newcommand{\parsection}[1]{\vspace{4pt}\noindent\textbf{#1.}}

\definecolor{peachorange}{RGB}{255,150,120}
\definecolor{warmgray}{RGB}{238,229,222}
\def\paperID{2338} 
\def\confName{ICCV}
\def\confYear{2025}

\title{GraspCoT: Integrating Physical Property Reasoning for 6-DoF Grasping under Flexible Language Instructions}
\author{Xiaomeng Chu$^1$, Jiajun Deng$^{2*}$, Guoliang You$^1$, Wei Liu$^1$, Xingchen Li$^1$, Jianmin Ji$^1$, Yanyong Zhang$^{1}$\thanks{Corresponding authors.} 
\\
\\
$^1$ University of Science and Technology of China,
$^2$ The University of Adelaide
}

\begin{document}
\maketitle
\begin{abstract}

Flexible instruction-guided 6-DoF grasping is a significant yet challenging task for real-world robotic systems.
Existing methods utilize the contextual understanding capabilities of the large language models (LLMs) to establish mappings between expressions and targets, allowing robots to comprehend users' intentions in the instructions.
However, the LLM's knowledge about objects' physical properties remains underexplored despite its tight relevance to grasping.
In this work, we propose \systemname, a 6-DoF grasp detection framework that integrates a Chain-of-Thought (CoT) reasoning mechanism oriented to physical properties, guided by auxiliary question-answering (QA) tasks. 
Particularly, we design a set of QA templates to enable hierarchical reasoning that includes three stages: target parsing, physical property analysis, and grasp action selection.
Moreover, \systemname presents a unified multimodal LLM architecture, which encodes multi-view observations of 3D scenes into 3D-aware visual tokens, and then jointly embeds these visual tokens with CoT-derived textual tokens within LLMs to generate grasp pose predictions.
Furthermore, we present \datasetname, a large-scale benchmark that fills the gap in public datasets for multi-object grasp detection under diverse and indirect verbal commands.
Extensive experiments on \datasetname demonstrate the superiority of our method, with additional validation in real-world robotic applications confirming its practicality.
The code is available at \href{https://github.com/cxmomo/GraspCoT}{https://github.com/cxmomo/GraspCoT}.

\end{abstract}

\section{Introduction}
\label{sec:intro}

\begin{figure}[t]
    \centering
    \includegraphics[width=0.96\columnwidth]{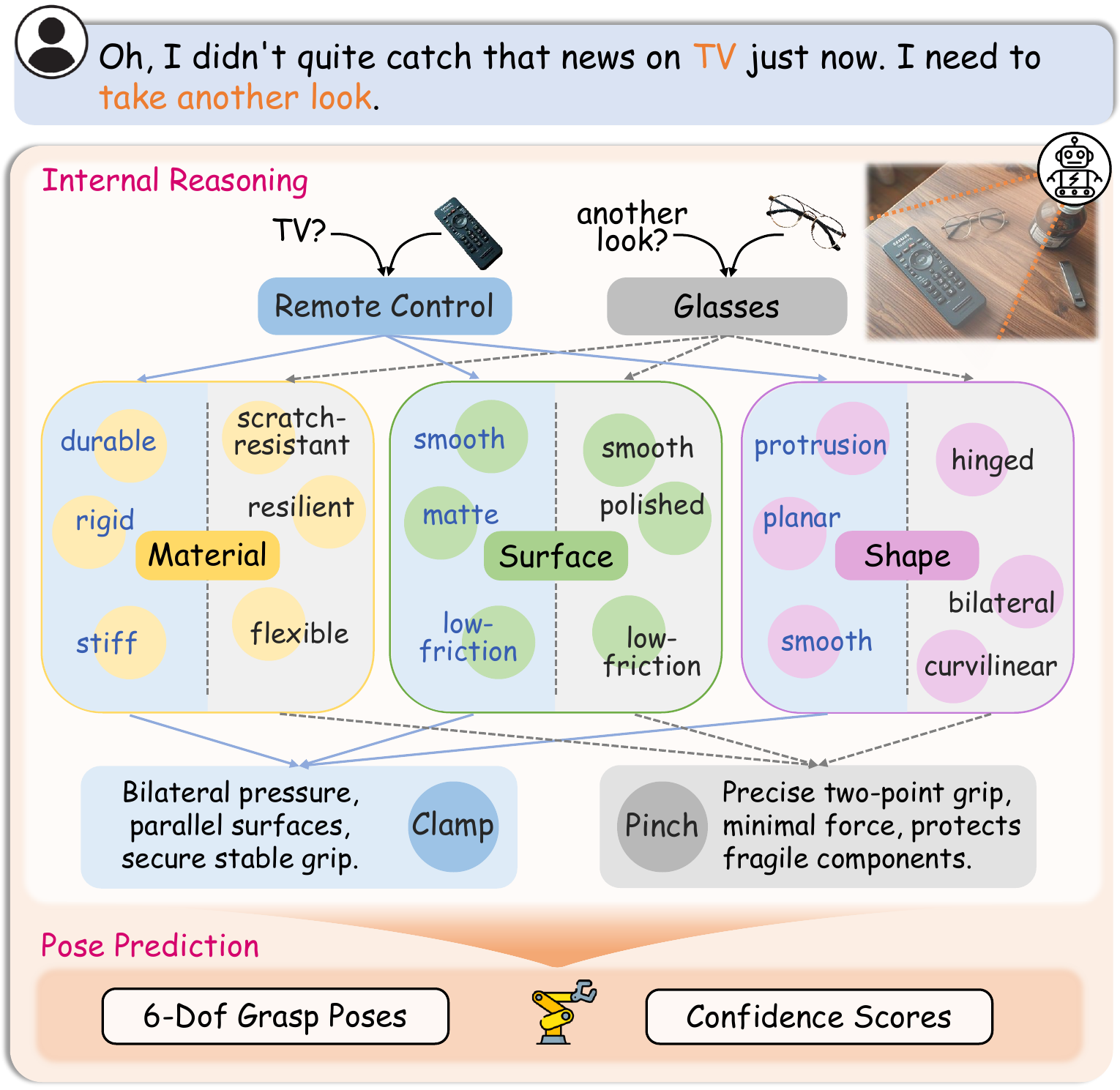}
    \vspace{-0.1cm}
    \caption{
    Example of \systemname. Facing a rewatching TV request, \systemname employs CoT reasoning, including target parsing, physical property analysis, and action selection, to interpret the scene and predict 6-DoF grasp poses with confidence scores for critical objects, namely remote control and glasses.
    }  
    \label{fig:example}
    \vspace{-0.2cm}
\end{figure}

The language-guided 6-DoF grasp detection task tackles the problem of robotic grasping that interprets human instructions and 3D environments~\cite{grasping_survey1,grasping_survey2,grasping_survey3}. This capability holds critical importance for domestic assistance, healthcare, and industrial automation. However, it presents greater complexity than its language-agnostic robotic grasping counterparts~\cite{schmidt2018grasping,kalashnikov2018scalable,morrison2018closing,le2010learning,Unigrasp}, mainly due to the cross-modal alignment between textual and visual representations.
Consequently, existing methods~\cite{3DAPNet,SunLFFX23,OVGNet} use a text encoder~\cite{CLIP,Bert,BLIP-2} to map textual and visual features into a unified latent space, enabling cross-modal correspondence for instruction-guided grasping.

With the emergence of LLMs and their multi-modal variants (MLLMs), there has been a surge of research interest in integrating semantic knowledge and common sense into object grasping. Early works leverage LLMs mainly for semantic understanding in task-oriented grasping. For example, GraspGPT~\cite{GraspGPT} uses LLMs to describe objects and task-relevant information, enabling interaction with novel object classes, while SemGrasp~\cite{SemGrasp} fine-tunes an MLLM to generate human-like grasps. These models primarily handle explicit instructions, such as ``Give me the remote control", where the target object and action are clearly stated. A more recent trend focuses on exploring the reasoning ability of LLMs to interpret implicit grasping instructions. 
The methods~\cite{ThinkGrasp,Reasoning_Grasping} following this trend infer the intended action from indirect cues, such as recognizing the need to fetch a remote control from ``I want to watch TV". 
Such practices makes human-computer interaction more natural by understanding users' intention.
Nevertheless, merely understanding the semantics or users' intentions does not fully leverage the common knowledge of LLMs to improve grasp detection. The physical properties of objects, essential for real-world manipulation, remain underexplored. 
Beyond geometric similarity, effective grasping requires reasoning about functional and physical attributes—for instance, a fragile vase demands precise neck grasping to prevent breakage, whereas a sturdy mug allows more flexible handling based on material strength and texture.

By consolidating this idea, we explore a new paradigm that can hierarchically integrate objects' physics-centric analyses into grasping by leveraging LLMs' remarkable reasoning capabilities. Besides, the previous methods generally assume that the grasping task only involves a single object, we give up such an assumption to make the grasping algorithm more applicable. Notably, in this work, we use ``\textit{flexible instruction}" to indicate the expression that does not explicitly specify the target object or even object quantity.

Formally, we introduce \textbf{\systemname}, a 6-DoF grasp detection framework featuring a physical-property-oriented chain-of-thought (CoT)~\cite{Chain-of-Thought} reasoning mechanism via auxiliary question-answering (QA) tasks. Specifically, we develop QA templates for a three-stage CoT pipeline covering target parsing, physical property analysis, and grasp action selection. 
QA texts and flexible language instructions are encoded as textual tokens.
Fig.~\ref{fig:example} provides an example of the CoT reasoning process of \systemname. 
Departing from conventional pipelines that decouple multimodal understanding into isolated modules, 
our framework integrates visual-textual comprehension directly into 6-DoF grasp pose prediction, inspired by breakthroughs in 3D MLLMs~\cite{LL3DA,3D-LLaVA,Ulip-2}. 
This unified architecture not only enhances cross-modal feature alignment but also reduces system complexity.
On the visual side, we construct multi-view observations of the 3D scene by projecting colored point clouds into complementary RGB images and depths. The images are processed into 2D patches via CLIP, enriched with depth-derived 3D position embeddings, and back-projected into geometrically grounded 3D tokens~\cite{LLaVA-3D,SpatialVLA,video_3dllm}. 
Finally, these 3D tokens, along with textual tokens mentioned above, undergo deep alignment within LLMs to predict grasp poses with confidence scores for the target objects.

Besides, there is no public benchmark for evaluating grasp detection under flexible instructions, especially for grasping multiple target objects. To bridge this gap, we propose \datasetname, a novel benchmark adapted from the Grasp-Anything-6D dataset~\cite{lg6d}. For each 3D scene, we employ structured prompt templates processed by Llama-3~\cite{LLaMA,llama_3} to generate linguistically diverse requirements that identify target objects through contextual clues rather than explicit identifiers. This benchmark challenges robotic systems to resolve referential ambiguity by joint visual perception with textual semantics, thereby comprehensively assessing their cross-modal cognitive abilities.

We evaluate \systemname on the \datasetname benchmark, demonstrating the superiority of our proposed \systemname for both pose estimation accuracy and collision avoidance rate.
Besides, we deploy the proposed \systemname on a Kinova Gen3 robotic arm. This extended experiment further validates its effectiveness in real-world robotic applications.

In summary, our main contributions are as follows:
\begin{itemize}
    \item We propose a physical-property-oriented CoT reasoning paradigm that bridges 3D scene understanding and grasp detection via auxiliary QA tasks. 
    \item We introduce a unified grasp detection framework, \systemname, that deeply embeds LLMs for the joint optimization of textual-visual feature alignment via dual-task learning. 
    \item We propose \datasetname, the first benchmark supporting multi-target grasping under flexible instructions. 
    \item We demonstrate the superior performance of \systemname by comprehensive evaluations on the \datasetname benchmark and real-world deployment on the robot arm. 
\end{itemize}

\section{Related Work}
\label{sec:related}

\begin{figure*}[t]
    \centering
    \includegraphics[width=0.96\textwidth]{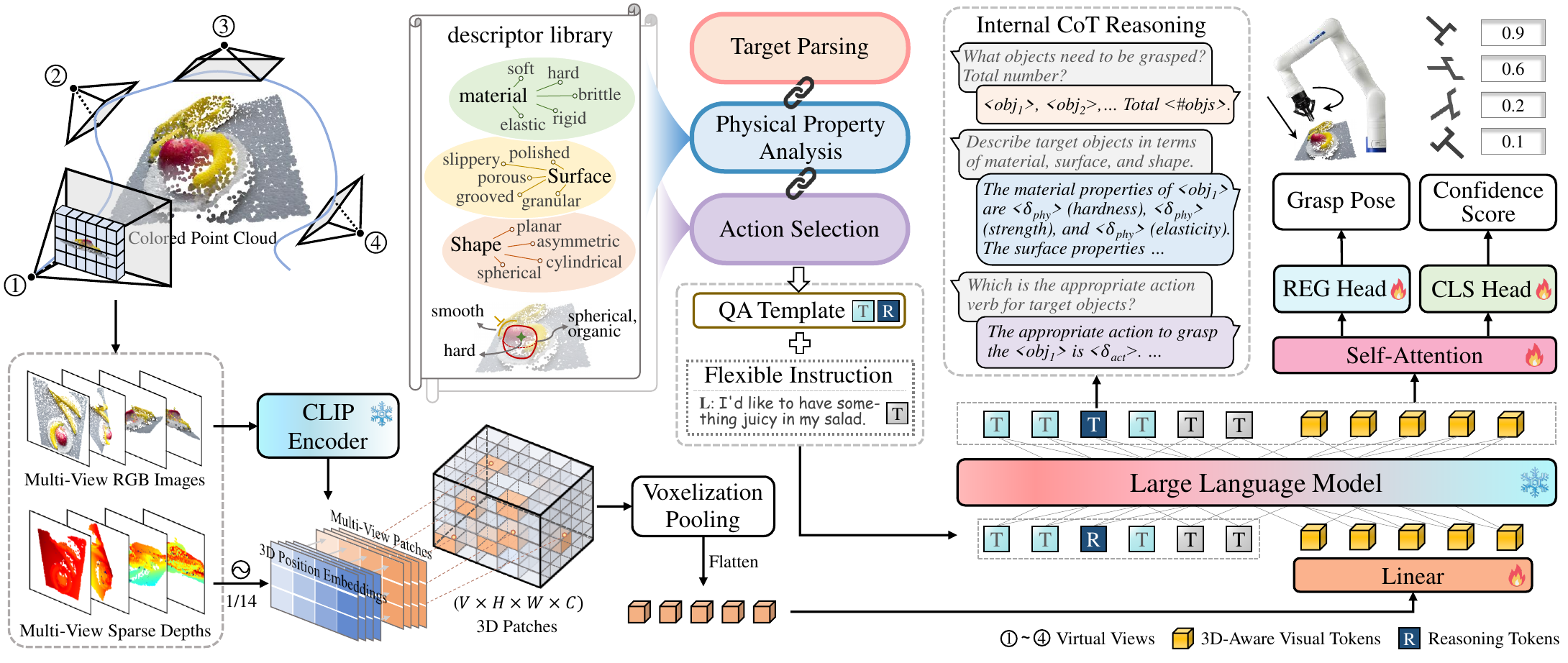}
    \vspace{-0.2cm}
    \caption{\textbf{Overview of \systemname.} 
    Given a colored point cloud, we project it into multi-view RGB-D images. CLIP-encoded RGB patches and depth-derived 3D positional embeddings are fused and back-projected into 3D patches, which are voxel-pooled and linearly projected into 3D-aware visual tokens. 
    We propose a three-stage Chain-of-Thought (CoT) framework based on fill-in-the-blank question-answering (QA) templates. This framework consists of three stages: target parsing, physical property analysis, and action selection, all enhanced with descriptor libraries. In this setup, the blank fields within the templates are represented as reasoning tokens, which the large language model (LLM) resolves into physical or action descriptors in order to generate answers. The fixed textual components of the templates serve as contextual QA tokens.
    After performing reasoning in the LLM, the autoregressive generated output tokens are all fed into a self-attention layer, followed by parallel grasp pose regression and confidence prediction.
    }
    \label{fig:arch}
    \vspace{-0.2cm}
\end{figure*}

\parsection{Robotic Grasping}
The emergence of grasping datasets~\cite{Graspnet-1billion,Acronym,Egad,Jacquard} lay the groundwork for data-driven grasping, while later works extend these foundations to diverse and task-aware scenarios.
6-DOF GraspNet~\cite{6dofgraspnet} employs variational autoencoders to sample diverse poses from partial observations, and AnyGrasp~\cite{Anygrasp} introduces temporal consistency for dynamic environments. Recognizing the need for hardware generalization, UniGrasp~\cite{Unigrasp} pioneers gripper-agnostic frameworks through kinematic latent encoding, while optimization-centric paradigms like SE(3)-DiffusionFields~\cite{SE3DiffusionFields} reformulate grasping as gradient-based pose-trajectory co-optimization. Recent efforts balance these directions: Wang et al.~\cite{Graspness} accelerates inference via geometric heuristics, and Wang et al.~\cite{Goalgrasping} enhances physical interaction through auxiliary task learning. Collectively, these works demonstrate a shift from isolated geometric analysis to hybrid systems that embed functional adaptability and computational efficiency within unified frameworks.

\parsection{Language-guided Grasping}
Recent advances in vision-language models have revolutionized language-guided robotic grasping by enabling semantic alignment between textual instructions and visual perception. 
Pioneering works~\cite{3DAPNet,lg6d,SunLFFX23,LLGD,OVGNet,ETOG,GLOVER,wei2024grasp} leverage a text backbone~\cite{CLIP,Bert} to embed textual features, where some explore diffusion models for 3D manipulation--3DAPNet~\cite{3DAPNet} extends grasp pose prediction paradigm to open-vocabulary affordance-pose joint learning,
while LGrasp6D~\cite{lg6d} introduces the Grasp-Anything-6D dataset and incorporates the negative prompt guidance for 6-DoF grasp detection.
OVGNet~\cite{OVGNet} proposes a unified visual-linguistic framework combining CLIP-style cross-modal alignment with grasp pose prediction, enabling generalization to unseen categories without category-specific retraining.
The emergence of (M)LLMs~\cite{LLaMA, LLaVA, BLIP-2} further expands the visual and textual interpretation capabilities in manipulation tasks~\cite{GraspGPT,SemGrasp,OWG,affordancellm}. 
ThinkGrasp~\cite{ThinkGrasp} introduces a language-guided, stepwise clutter removal strategy powered by GPT-4o's contextual inferring, enabling robotic grasping in occluded environments through goal-oriented obstacle displacement and few-step target recovery.
RT-Grasp~\cite{RT-Grasp} proposes a framework that unlocks LLMs' capacity for numerical prediction in robotics by embedding structured inferring phases into training, while Jin et al.~\cite{Reasoning_Grasping} leverage an end-to-end multimodal LLM framework to bridge implicit human intentions to grasp poses, enabling object/part-level grasping under indirect instructions.
Recent efforts combine visual grounding with attribute prediction of required objects using multimodal foundation models.
ShapeGrasp~\cite{ShapeGrasp} pioneers zero-shot task-oriented grasping through geometric decomposition of objects into convex shapes represented as attribute-rich graphs, combined with LLM-driven semantic reasoning to infer part-task utility.
DeliGrasp~\cite{DeliGrasp} proposes LLM-driven object properties prediction for adaptive grasping, converting semantic object descriptions into grasp policies via inferred mass/friction/spring parameters.
Instead of requiring LLMs to directly predict precise physical parameters--a task challenging current LLMs' capabilities--our method encodes physical properties into structured descriptors, explicitly guiding grasp pose prediction while avoiding error-prone low-level physics estimation.

\section{Method}
\label{sec:method}

In this section, we formally define the task of flexible-instruction-guided 6-DoF grasp detection, which requires jointly interpreting visual scene understanding and flexible language to determine optimal grasp configurations. 
Next, we present the framework, detailing its core components: the physical-property-oriented Chain-of-Thought (CoT) design and grasp decoding with the auxiliary question-answering (QA) task via large language models (LLMs). Finally, we outline the construction of the \datasetname benchmark, designed to evaluate grasping under linguistically flexible instructions.

\subsection{Preliminary}

The flexible-instruction-guided 6-DoF grasp pose detection takes as input a colored point cloud $\mathcal{P} = $\{$ (p_i, c_i) $\}$_{i=1}^N$, where $p_i \in \mathbb{R}^3$ represents the 3D coordinates and $c_i \in \mathbb{R}^3$ denotes the RGB color of each point, along with a flexible language instruction $t$. The output consists of CoT-relevant questions and answers $\mathcal{Q} = $\{$ (q_k, a_k) $\}$_{k=1}^M$ that analyze the categories, physical properties, and grasp action of the target object(s), followed by the generation of multiple 6-DoF grasp poses $g \in SE(3)$ for the identified target(s).
Formally, this process $\Phi$ is defined as:  
\begin{equation}
\centering
\Phi: (\mathcal{P}, t) \mapsto (\mathcal{Q}_\text{CoT}, \mathcal{G}),
\end{equation}
where $\mathcal{G}$ represents the grasp pose set. Each $g_j = (R, T, w) \in \text{SE}(3)$ defines a 6-DoF gripper configuration composed of the orientation $R \in \text{SO}(3)$, the position $T \in \mathbb{R}^3$, and the suitable gripper width $w \in \mathbb{R}^+$.

\begin{figure}[t]
    \centering
    \includegraphics[width=0.88\columnwidth]{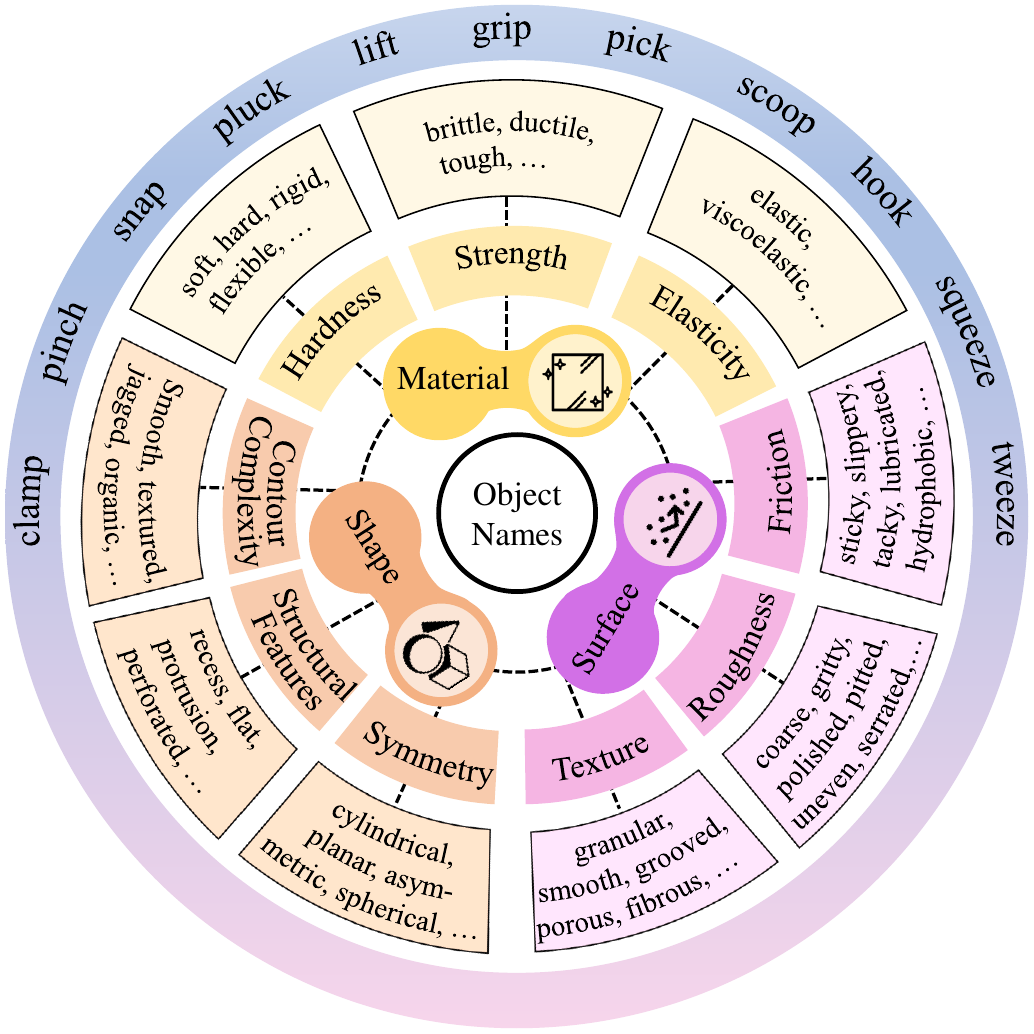}
    \vspace{-0.1cm}
    \caption{
    Hierarchical CoT Reasoning: target parsing (inner ring), multi-aspect physical property analysis (material/surface/shape in middle three rings), and grasp action selection represented by the action descriptor libraries (outer ring).
    }  
    \label{fig:physical_cot}
    \vspace{-0.3cm}
\end{figure}

\subsection{Model Architecture}

With growing advances in 3D multi-modal LLMs (MLLMs)~\cite{3D-LLM,Agent3D-Zero,Scene-LLM,Grounded_3D-LLM}, recent breakthroughs have been achieved in aligning 3D geometric representations with linguistic instructions. 
As shown in Fig.~\ref{fig:arch}, our architecture operates on colored point cloud inputs through a dual-branch processing pipeline. 
In the visual processing branch, we construct multiple visual viewpoints for projection to generate multi-view sparse RGB-D representations.
These projections undergo CLIP-based feature encoding, which is subsequently combined with 3D position embeddings from depth data. After back-projection, voxel pooling, and linear projection, we transform the reconstructed 3D patches into 3D-aware visual tokens. Simultaneously, the textual branch employs a hierarchical CoT reasoning framework comprising target parsing, physical property analysis, and action selection. To guide a robust reasoning process, we establish open-world descriptor repositories and develop semi-supervised QA templates that produce supervised QA tokens along with unsupervised reasoning tokens for descriptor instantiation. 
The autoregressive tokens produced by the LLM’s reasoning stage are first passed into a self-attention layer, then jointly utilized for parallel predictions of grasp poses and their confidence scores.

\subsection{Physical-Property-Oriented CoT Design}
\label{subsec:graspqa}

To bridge scene understanding and grasp poses, we design fill-in-the-blank QA templates and implement a three-stage CoT reasoning pipeline that sequentially grounds target categories, infers physical properties, and maps grasp actions.
Structured QA templates are designed as follows to guide the reasoning process for grasp detection:  
\begin{itemize}[leftmargin=22pt, noitemsep]
    \item[\textit{Q1}:] \textbf{Target Parsing} - Which objects need to be grasped?
    \item[\textit{A1}:] [\texttt{<$\texttt{obj}_1$>}, \texttt{<$\texttt{obj}_2$>}, ...]. Total \texttt{<\#objs>}.
    \item[\textit{Q2}:] \textbf{Physical Property Analysis} (e.g., material) - For each target object, analyze its material from three aspects: hardness [options: soft, hard, rigid, flexible, etc.], strength [options: brittle, ductile, tough, etc.], and elasticity [options: elastic, viscoelastic, etc.].
    \item[\textit{A2}:] The material properties of \texttt{<$\texttt{obj}_\texttt{k}$>} are \texttt{<$\delta_{\texttt{phy}}$>} (hardness), \texttt{<$\delta_{\texttt{phy}}$>} (strength), and \texttt{<$\delta_{\texttt{phy}}$>} (elasticity).
    \item[\textit{Q3}:] \textbf{Grasp Action Selection} - For each target object, select an appropriate grasp action [options: clamp, pinch, snap, pluck, lift, grip, etc.].
    \item[\textit{A3}:] The appropriate verb to grasp the \texttt{<$\texttt{obj}_\texttt{k}$>} is \texttt{<$\delta_{\texttt{act}}$>}.
\end{itemize}

First, the target parsing stage resolves referential ambiguities in an answering template with keyword slots, where target categories are explicitly supervised. 
Next, the pipeline takes advantage of the common sense and reasoning ability of MLLMs to infer physical properties, which are hierarchically categorized into three primary groups: material, surface, and shape\footnote{While ``shape" is geometrically defined, our analysis targets inferring mass distribution and optimal grasping contact forces, thus integrating it into the broader category of physical properties for unified reasoning.}.
Each property is further subdivided into three fine-grained characteristics described by appropriate physical descriptors $\delta_{phy}$ in open-world libraries, as illustrated in Fig.~\ref{fig:physical_cot}. 
For example, a glass cup’s material properties are characterized as \texttt{<rigid>} in hardness, \texttt{<brittle>} in strength, and \texttt{<inflexible>} in elasticity. Similarly, its surface attributes are described as \texttt{<smooth>} in texture, \texttt{<polished>} in roughness, and \texttt{<slippery>} in friction. This structured classification enables a comprehensive representation of physical properties for the prediction of grasp poses.
Finally, the action selection stage requires LLMs to select reasonable action descriptors $\delta_{act}$ (e.g., ``pinch" for fragile objects, ``clamp" for textured surfaces) from a predefined verb library, dynamically linking these choices to geometric parameters such as gripper orientation and contact force. 
Notably, the physical/action descriptor tokens remain unsupervised, leveraging LLMs' extensive corpora knowledge to bypass manual annotation of exhaustive object-specific properties.

\subsection{Auxiliary QA Task and Grasp Decoding}

The synergistic integration of 3D-aware visual tokens and flexible-instruction tokens enhances the model's comprehension of 3D environments and interactive intent. 
As mentioned before, \systemname unifies grasp-related QA tasks and grasp pose prediction through joint training, enabling deep fusion of task-relevant token representations within the pre-trained LLM architecture.
We implement answer templates for three question categories in Sec.~\ref{subsec:graspqa}.
Crucially, the physical properties of targets and grasp actions remain unspecified descriptors that require no explicit annotation. Instead, we introduce special ``reasoning tokens", which are resolved into physical or action descriptors by the LLM, to represent these latent variables. During training, these tokens receive masked supervision with zero loss weighting.

The core output of \systemname consists of multiple 6-DoF grasp poses with corresponding confidence scores, which inspired us to adopt a query-based detection decoding approach. Specifically, we first process all output tokens from the LLM through a self-attention layer, then feed them into two separate heads: a regression head for grasp pose estimation and a classification head for confidence prediction. During training, we employ the Hungarian matching algorithm to align predicted grasp poses with ground truth annotations and compute L1 loss $L_{\text{Reg}}$ for pose refinement. Simultaneously, we apply focal loss~\cite{focalloss} $L_{\text{Cls}}$ to optimize the confidence scores of the predicted poses.
Combined with the cross-entropy loss $L_{\text{QA}}$ for the QA task, the total loss of the model integrates three components as follows:
\begin{equation}
L = L_{\text{QA}} +  L_{\text{Reg}} +  L_{\text{Cls}}.
\end{equation}

\subsection{\datasetname Benchmark Construction}

\begin{table*}[t]
  \centering
  \small
    \setlength\tabcolsep{19pt}
    \begin{tabular}{l|ccccc}
    \toprule
      Methods &  CR@0.4$\uparrow$ & CR@0.3$\uparrow$  & CR@0.2$\uparrow$ &  EMD$\downarrow$ & EW-CFR$\uparrow$  \\
    \midrule
     3DAPNet~\cite{3DAPNet} &  0.9249 & 0.7378  &  0.2904  &  0.3066 & 0.3521   \\
     LGrasp6D~\cite{lg6d} & 0.9294  & 0.7567  & 0.3349  & 0.2935  & 0.3847  \\
     \midrule
     \systemname (ours) w/o CoT & \underline{0.9468} &  \underline{0.7941} &  \underline{0.4164}  &  \underline{0.2878}  & \underline{0.4173}    \\
     \systemname (ours) & \textbf{0.9715}  &  \textbf{0.8797} &  \textbf{0.5587}  &\textbf{ 0.2520}  & \textbf{0.4229}  \\
    \bottomrule
    \end{tabular}
    \vspace{-0.2cm}
    \caption{Results on \datasetname benchmark under flexible language instructions.
    ``CR@$\theta$" denotes the success rate when predicted poses are within a threshold $\theta$ of ground truth poses.
    }
    \vspace{-0.3cm}
  \label{tab:benchmark_results}
\end{table*}

Most existing language-guided grasping benchmarks utilize explicit instructions such as ``Bring me a bottle of water" or ``Grab the black remote control", which explicitly specify target objects by name and salient visual attributes. However, real-world human-robot interaction demands support for implicit requests. For instance, a user stating ``I’m thirsty" or ``I want to watch TV", where the robot should infer the need for water or the remote control. 
Equipped with visual scene understanding, the robot can further engage in contextual reasoning: upon detecting ``glasses" in the environment while addressing a ``watch TV” request, it might proactively retrieve them for the user. 
We define such commands as flexible instructions, where one or more target objects are required but neither explicitly named nor uniquely defined. The robot must interpret the user’s intent through contextual cues within a plausible scenario, potentially retrieving one or multiple contextually relevant targets.

\begin{figure}[t]
    \centering
    \includegraphics[width=0.95\columnwidth]{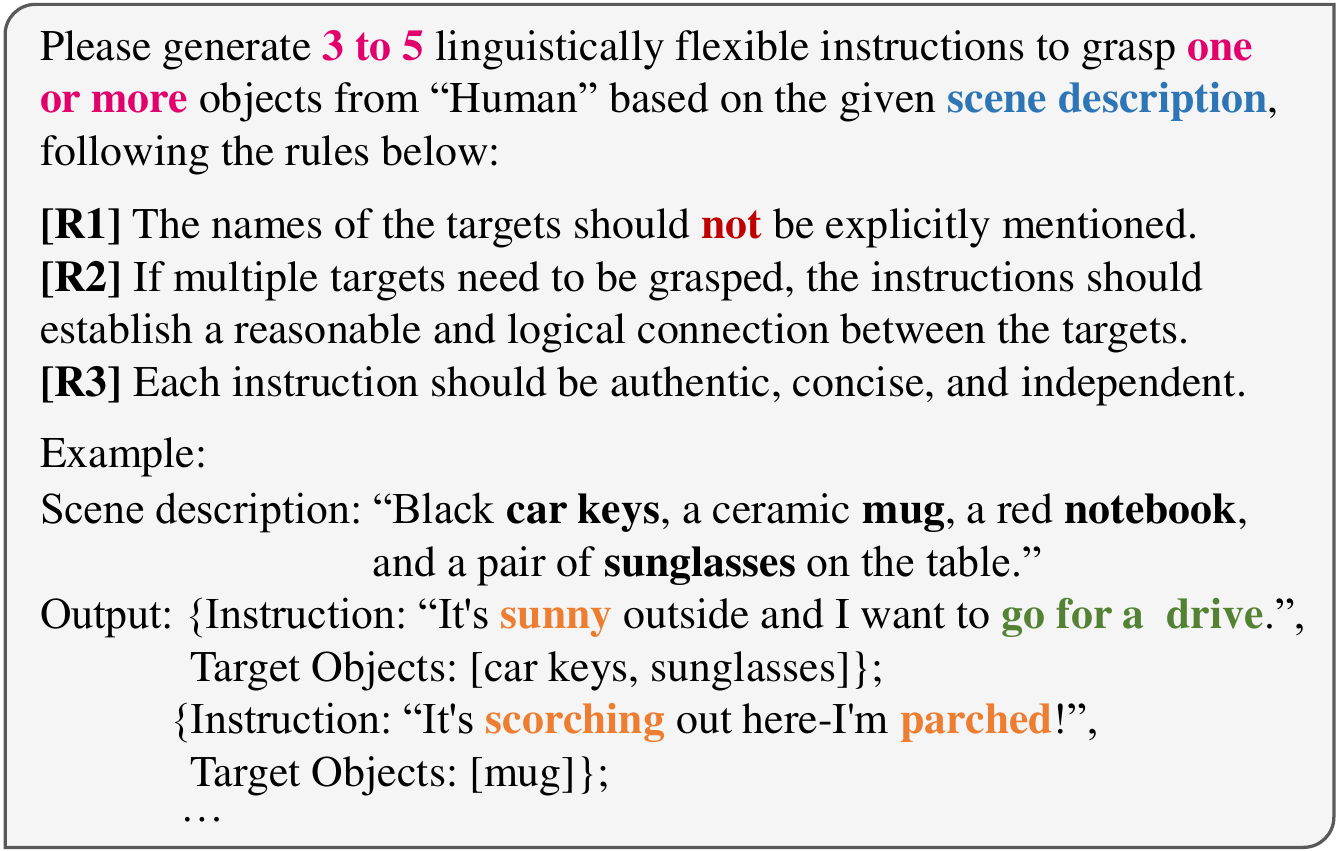}
    \vspace{-0.2cm}
    \caption{
    Prompt used for flexible instruction generation.
    }  
    \label{fig:prompt}
    \vspace{-0.3cm}
\end{figure}

To address the lack of public benchmarks supporting multi-target grasp detection under such flexible instructions,
we develop \datasetname, a benchmark containing one million (1M) 3D colored point cloud scenes with approximately 3M objects, sourced from the Grasp-Anything-6D dataset~\cite{lg6d}, which provides scene descriptions, explicit instructions, and grasp pose annotations. For each scene, we generate 3-5 flexible instructions using the open-source LLM Llama3-70B~\cite{LLaMA} guided by a customized prompt template. As shown in Fig.~\ref{fig:prompt}, the prompt directs the LLM to create plausible requests that implicitly demand the grasping of one or multiple objects without naming targets. For example, in a scene containing ``black car keys, a ceramic mug, a red notebook, and sunglasses on the table", the Grasp-Anything-6D dataset provides explicit instructions like ``Grasp the black car keys", whereas our method generates contextual statements such as ``It’s sunny outside and I want to go for a drive", implicitly requiring the robot to retrieve both car keys and sunglasses.
\datasetname fills the critical gap in flexible-instruction-driven multi-target grasping benchmarks. Its generation design can be further integrated with detection/segmentation models~\cite{sam, UniDetector} to extend to any other grasping dataset.

\section{Experiments}
\label{sec:experiments}

In this section, we validate the effectiveness of \systemname using flexible instructions on both the \datasetname benchmark and real-world experimental platforms.
\subsection{Main Results on \datasetname Benchmark}
\parsection{Setup}
To evaluate the effectiveness of our grasp pose prediction method, we employ three metrics: coverage rate (CR)~\cite{6dofgraspnet}, earth mover’s distance (EMD)~\cite{SE3DiffusionFields}, and collision-free rate (CFR)~\cite{regnet} with appropriate adaptations.
CR quantifies the proportion of ground-truth grasp poses that are sufficiently covered by predictions. Specifically, it measures the percentage of ground-truth poses where at least one predicted pose lies within a distance threshold $\theta$ (Euclidean distance in SE(3) space). We evaluate CR under three increasingly stringent thresholds: $\theta$ = 0.4, 0.3, and 0.2.
EMD measures how holistically the predicted poses match the spatial distribution of ground-truth annotations in SE(3) space, where lower values indicate better distributional consistency.
CFR evaluates the physical feasibility of predicted grasp poses by measuring their ability to avoid collisions with the target object and environment. 
However, a critical limitation arises when directly using the CFR metric: erroneous grasp poses far from the target object may artificially achieve higher scores because they avoid contact with environmental point clouds entirely. 
To jointly evaluates geometric consistency and physical feasibility where higher values indicate safer grasps, we propose an EMD-Weighted CFR (EW-CFR) metric defined as follows:
\vspace{-0.1cm}
\begin{equation}
    \small
    \centering
    \text{EW-CFR}= \frac{1}{N}\sum_{i=1}^N \big(1 - \mathcal{C}(\mathcal{P}, \mathcal{M}(g_i))\big) \cdot \big(1 - \frac{\text{EMD}(g_i, \mathcal{\hat{G}})}{R}\big), 
\vspace{-0.1cm}
\end{equation}
where $R$ normalizes the EMD($g_i$, $\mathcal{\hat{G}}$). 
The function $\mathcal{C}$ denotes a binary collision check between the point cloud $\mathcal{P}$ and the gripper model $\mathcal{M}(g_i)$. 

The \datasetname dataset is partitioned following conventional practice, with 80\% of samples dedicated to training and the remaining 20\% preserved as the evaluation set.

\parsection{Baseline}
In comparative experiments, we benchmark our method against two recent language-guided 6-DoF grasp detection models: 3DAPNet~\cite{3DAPNet} and LGrasp6D~\cite{lg6d}. 
To align with scenarios under flexible instructions, we replace their original explicit grasp command inputs with flexible instructions while preserving their default training protocols.
Specifically, we eliminate 3DAPNet's affordance prediction module to better align with our task requirements.
For consistency across all models, the valid grasp pose's positions (x, y, z) and orientations along x- and y-axes are limited to the range [-1, 1], and z-axis orientations are bounded within [0, $\pi$]. 
To ensure fairness, we uniformly sample 600 valid grasp poses of each scene generated by 3DAPNet and LGrasp6D for validation. Since each grasp pose in \systemname is assigned a confidence score, we select the top 600 highest-confidence poses as valid candidates.
\begin{figure*}[t]
    \centering
    \includegraphics[width=0.92\textwidth]{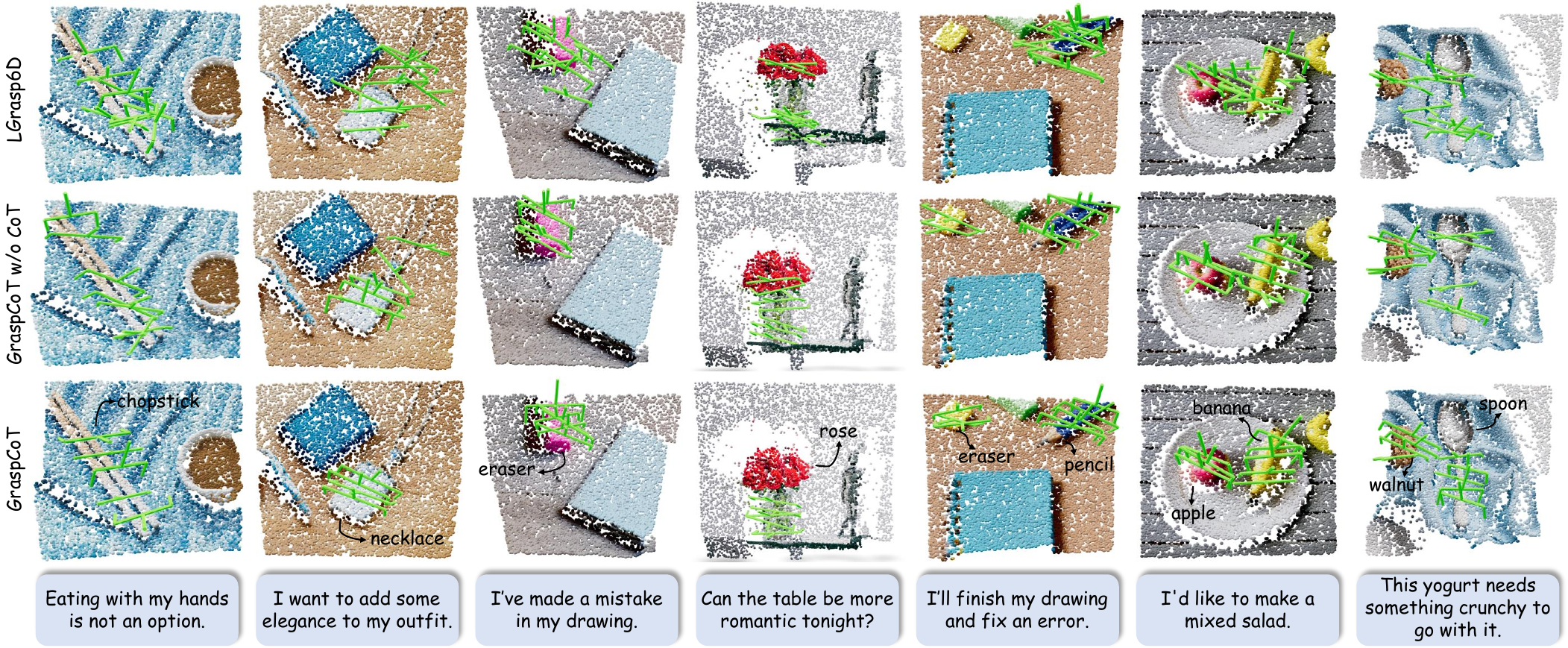}
    \vspace{-0.2cm}
    \caption{Visualization of 6-DoF grasp detection results based on flexible instructions.
    }
    \label{fig:benchmark_vis}
    \vspace{-0.3cm}
\end{figure*}

\parsection{Grasp Label Pruning}
Wu~\cite{Economic_Grasp} et al. demonstrate that excessively dense supervision incurs label ambiguity, particularly caused by the diversity of grasp orientations, as a critical bottleneck hindering model convergence.
Our analysis reveals that certain objects in the Grasp-Anything-6D dataset contain over 700 annotated grasp poses.  
This overabundance of positive samples not only exacerbates the aforementioned issues but also severely challenges Hungarian matching during training.
Consequently, we implemented a pruning strategy in \datasetname to reduce similar grasp labels, especially redundant rotation-similar annotations, per object to 100 instances. The detailed pruning method is provided in the supplementary material.

\parsection{Training}
Our models are trained using the Adam optimizer with a batch size of 8 per GPU over 4 epochs, initialized with LLaVA-3D~\cite{LLaVA-3D} pre-trained weights from the HuggingFace repository. The learning rate followed a cosine annealing schedule, starting at 1$\times10^{-4}$. All experiments are executed on 8$\times$24GB NVIDIA RTX3090 GPUs.

\begin{table}[t]
  \centering
  \small
    \setlength\tabcolsep{2.5pt}
    \begin{tabular}{cccc|ccc}
    \toprule
    Material & Surface & Shape & Action & CR@0.3$\uparrow$  &  CR@0.2$\uparrow$ &  EMD$\downarrow$   \\
    \midrule
      & & &  &  0.7941 &  0.4164  &  0.2878   \\
     \checkmark & & & &  0.8507 &  0.5091 & 0.2698   \\
      & \checkmark &  & & 0.8334  & 0.4818 & 0.2738 \\
      & &\checkmark  & & 0.8306 &  0.4716 & 0.2742 \\
      \checkmark& \checkmark& & & 0.8609 &  0.5177 & 0.2653   \\
      \checkmark& \checkmark& \checkmark& &  \underline{0.8717} & \underline{0.5386}   & \underline{0.2572}  \\ 
     \checkmark & \checkmark&\checkmark &  \checkmark &  \textbf{0.8797} &  \textbf{0.5587}  & \textbf{0.2520}   \\
    \bottomrule
    \end{tabular}
    \vspace{-0.2cm}
    \caption{
    Ablation study on physical property analysis and grasp action selection in CoT Reasoning.
    }
    \vspace{-0.3cm}
  \label{tab:ablan_cot}
\end{table}

\parsection{Quantitative Results}
In Tab.~\ref{tab:benchmark_results}, our method presents superior performance across all metrics on the \datasetname benchmark under flexible language instructions. Notably, \systemname attains a CR of 0.5587 at the strictest threshold ($\theta$=0.2), outperforming 3DAPNet and LGrasp6D by a large margin of 0.2683 and 0.2238, respectively. This substantial improvement highlights our method’s ability to generate geometrically precise grasp poses even under high-precision requirements. The progressive performance gap across $\theta$=0.4, 0.3, and 0.2 further demonstrates that our approach maintains effectiveness as task difficulty escalates.
The EMD of \systemname reflects a 14.1\% relative reduction compared to LGrasp6D, indicating superior alignment between the predicted and ground-truth pose distributions.
For our proposed EW-CFR metric, \systemname achieves a score of 0.4229, surpassing 3DAPNet and LGrasp6D by 20.1\% and 9.9\%, respectively. 
The w/o CoT version of \systemname confirms the critical role of our CoT reasoning pipeline: removing CoT results in a 25.4\% decline in CR@0.2 and a 9.1\% increase in EMD, respectively.

\parsection{Qualitative Results}
In Fig.~\ref{fig:benchmark_vis}, we present the comparative visualization results of \systemname for 6-DoF grasp detection under flexible instructions, covering both single- and multi-target scenarios. 
Our method exhibits more precise and contextually reasonable predictions in target parsing and grasp positioning. For example, \systemname correctly identifies grasp regions such as the body of a rose vase or the handle of a spoon. Notably, \systemname and the w/o CoT version both effectively avoid target omission in multi-target tasks, demonstrating its enhanced comprehension of both linguistic instructions and 3D scene contexts.

\begin{table}[t]
  \centering
  \small
    \setlength\tabcolsep{4.5pt}
    \begin{tabular}{l|c|cccc}
    \toprule
      \#Targets & CoT & CR@0.4$\uparrow$ & CR@0.3$\uparrow$  & CR@0.2$\uparrow$ &  EMD$\downarrow$  \\
    \midrule
     Single &  & 0.9465 &  0.7896  &  0.4108  & 0.2810     \\
     Multiple &  & 0.9492 &  0.7977 &  0.4183 & 0.3106   \\
     \midrule
     Single  & \checkmark & 0.9723 & 0.8840 & 0.5590   & 0.2453     \\
     Multiple  & \checkmark &  0.9680 & 0.8754  &  0.5571  &  0.2756     \\
    \bottomrule
    \end{tabular}
    \vspace{-0.2cm}
    \caption{
    Grasp detection results of \systemname vs. the w/o CoT version in single- and multi-target scenarios.
    }
    \vspace{-0.3cm}
  \label{tab:ablan_single_multi}
\end{table}

\subsection{Ablation Studies}

\parsection{Physical-Property-Oriented CoT Reasoning}
As shown in Tab.~\ref{tab:benchmark_results}, CoT reasoning significantly enhances \systemname's performance, validating its necessity. Tab.~\ref{tab:ablan_cot} quantifies the contributions of material, surface, and shape property analysis within the CoT pipeline, alongside grasp action selection. Material property analysis notably elevates performance, improving CR@0.3 and CR@0.2 by 7.1\% and 22.3\% respectively over the w/o CoT version, while reducing EMD by 6.3\%, demonstrating its critical role in refining pose prediction. Integrating surface or shape property sorely also boosts performance, achieving a 15.7\% and 13.3\% CR@0.2 gain, respectively. 
This finding reveals that even single-aspect physical property integration exerts a significant directional influence on LLMs' decoding of grasp-relevant hidden states.
Cumulative integration of all three physical analyses achieves the second-best pose estimation quality, reaching an EMD of 0.2572. Grasp action selection further enhances precision, resulting in a 3.7\% increase in CR@0.2 and a 2.0\% reduction in EMD.

\parsection{Single- and Multi-Target Grasping}
Our default evaluation protocol randomly selects one flexible instruction from 3–5 variants per scene, encompassing both single- and multi-target grasping scenarios. 
We then detail the respective performance of \systemname and its w/o CoT version in these two scenarios. 
In Tab.~\ref{tab:ablan_single_multi}, the w/o CoT version shows marginally superior CR metrics for multi-target grasping compared to single-target cases, albeit at the cost of 10.5\% higher EMD. When integrating CoT reasoning, significant improvements emerge: single-target performance achieves 12.0\% improvement in CR@0.3, 36.1\% improvement in CR@0.2, a 12.7\% decrease in EMD, while multi-target grasping attains 9.7\% improvement in CR@0.3, 33.2\% improvement in CR@0.2, and 11.3\% decrease in EMD. These results demonstrate that our CoT mechanism delivers stronger optimization for single-target scenarios because physical property analysis targets the object itself.

\parsection{Number of Virtual Views}
As presented in Tab.~\ref{tab:num_view}, we evaluate the impact of varying numbers of complementary virtual views, up to five. 
In ablation studies with no fewer than four views, performance using k (\textless4) views reveals that CR@0.2 experiences the most significant fluctuations: a 3.70\% relative increase when using 3 versus 2 views, followed by a substantial 12.6\% leap at 4 views.
Beyond four views,  performance plateaus indicate limited enhancement. We therefore adopt four complementary views as the default configuration, optimally balancing precision gains against computational overhead.

\subsection{Real-World Experiment}

\begin{figure}[t]
\small
\centering
\begin{tabular}{c}
\includegraphics[width=0.9\columnwidth]{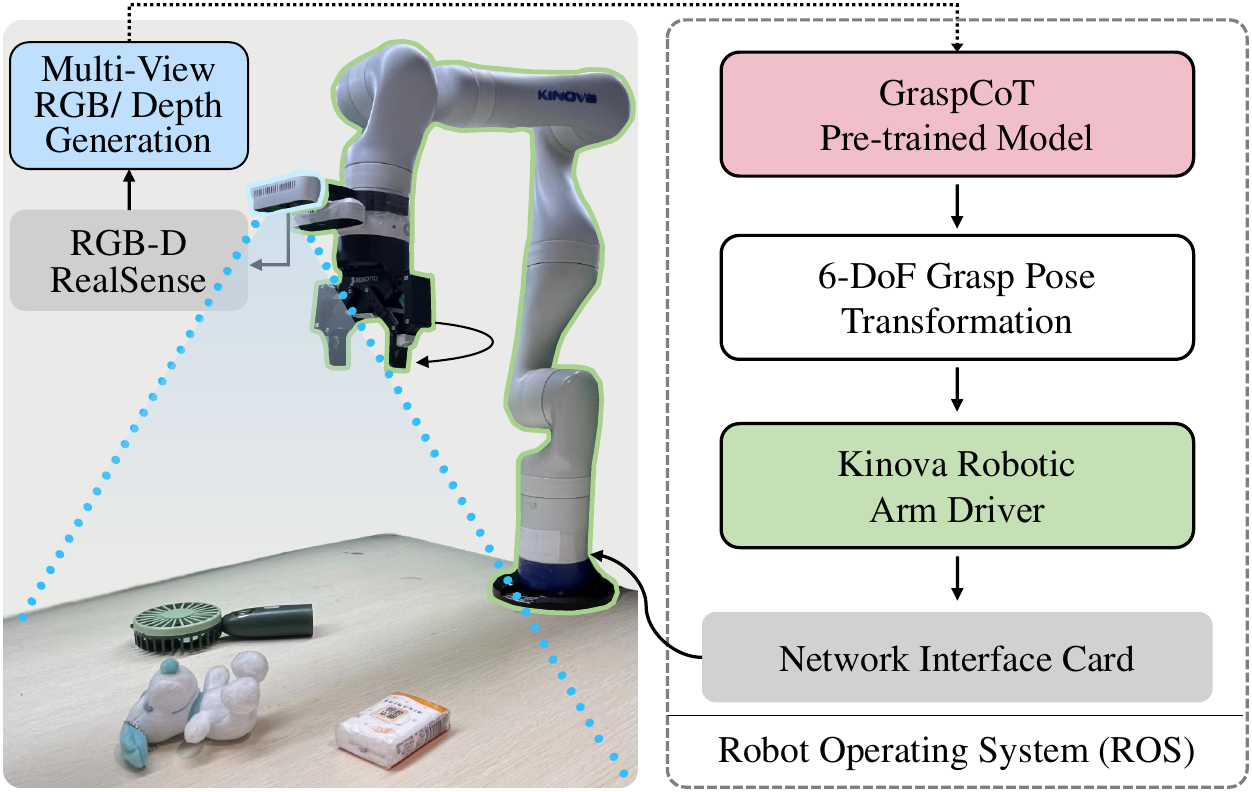} \\
(a) Real-world experimental setup. \\
\includegraphics[width=0.9\columnwidth]{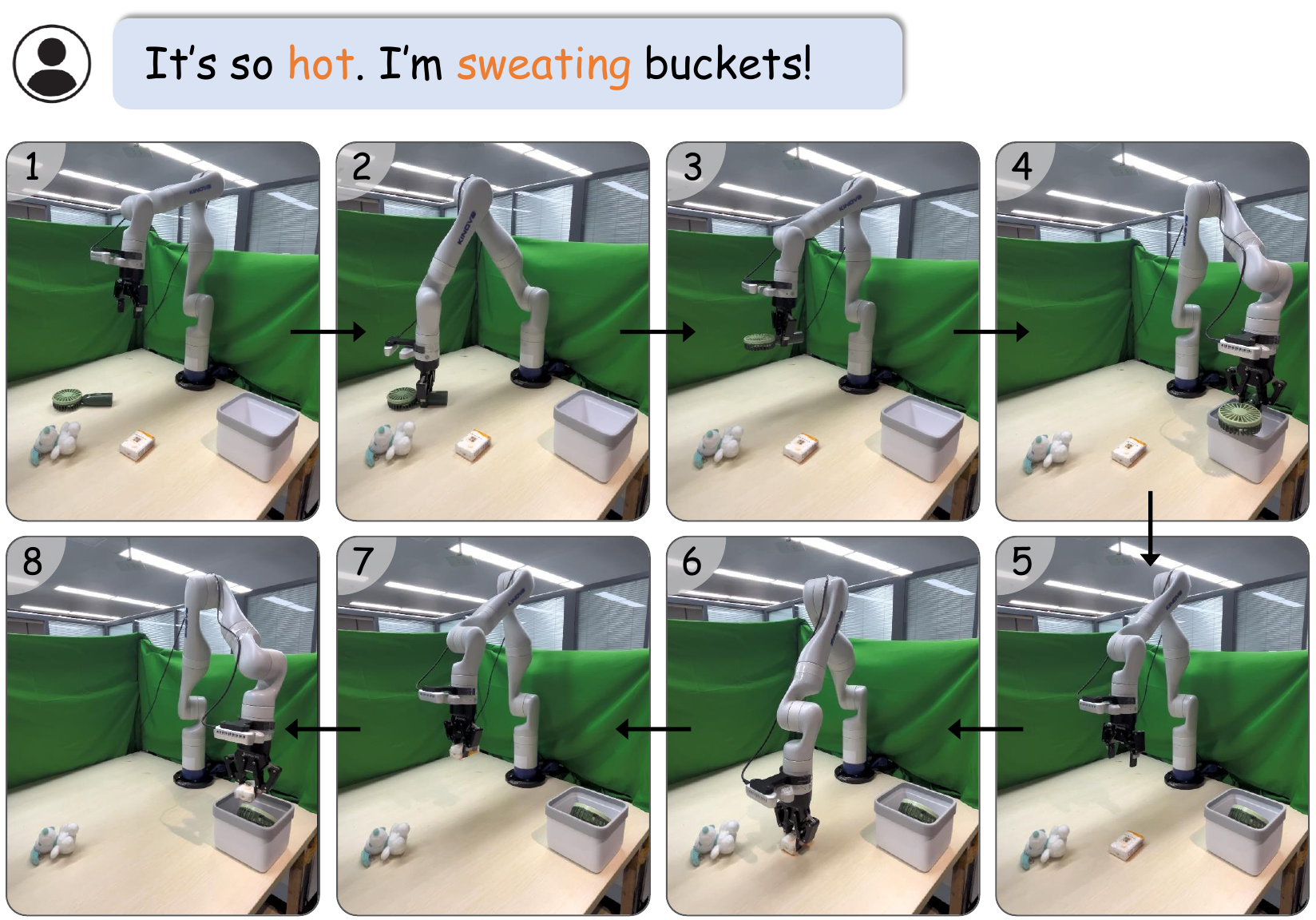} \\
(b) Example of the grasping task. \\
\end{tabular}
\vspace{-0.2cm}
\caption{An application example on a Kinova Gen3 robotic arm. The video demonstration is included in supplementary materials.}
\label{fig:real_world}
\vspace{-0.3cm}
\end{figure}

\parsection{Setup}
We deploy a Kinova Gen3 robotic arm with an Intel RealSense D435i depth camera, as shown in Fig.~\ref{fig:real_world}. The system detects 6-DoF grasp poses from RGB-D data, maps them to end-effector poses via hand-eye calibration, and executes trajectories autonomously. Our ROS-based platform (NVIDIA RTX 3090) processes real-time camera data and controls the robot via a network interface.
We categorize the test scenarios into explicit and flexible instructions, further dividing the latter into single-target grasping and multi-target sequential grasping. For each setting, we conduct 35 repeated grasping experiments, with the success rate as the evaluation metric.
Partial successes (e.g., grasping one of two required targets) are counted as individual successes, with the rate defined as successful grasps / total targets.

\parsection{Results}
As shown in Tab.~\ref{tab:real_world}, when handling explicit grasping instructions, CoT-enhanced \systemname achieved a 54.2\% success rate, outperforming its non-CoT counterpart by 2.8\%. For flexible instructions, the CoT version maintained stable performance with a 5.6\% advantage over the non-CoT variant, further validating its advantages in optimizing the grasp pose targeted at the object itself. 
Notably, in multi-target grasping scenarios where avoiding omissions of targets with weaker implicit relevance remains challenging, the CoT implementation improved success rates from 44.0\% to 46.7\%, which is a 2.7\% enhancement, verifying its effectiveness in handling complex requirements.

\begin{table}[t]
  \centering
  \small
    \setlength\tabcolsep{3pt}
    \begin{tabular}{c|ccccc}
    \toprule
      \#Views &  CR@0.4$\uparrow$ & CR@0.3$\uparrow$  & CR@0.2$\uparrow$ &  EMD$\downarrow$ & EW-CFR$\uparrow$ \\
    \midrule
     2 & 0.9637  &  0.8485   &  0.4785 &  0.2600  &  0.3984 \\
     3  &  0.9666  &  0.8566  & 0.4962  & 0.2569   & 0.4103   \\
     4  & \textbf{0.9715} & \underline{0.8797} & \textbf{0.5587} & \underline{0.2520} & \underline{0.4229}  \\
     5  & \underline{0.9700} & \textbf{0.8811} & \underline{0.5581}  & \textbf{0.2517} & \textbf{0.4239} \\
    \bottomrule
    \end{tabular}
    \vspace{-0.2cm}
    \caption{
    Performance analysis of virtual view quantity.
    }
    \vspace{-0.3cm}
  \label{tab:num_view}
\end{table}

\begin{table}[t]
  \centering
  \small
    \setlength\tabcolsep{14pt}
    \begin{tabular}{l|c|cc}
    \toprule 
      \multirow{2}{*}{Methods} &  \multirow{2}{*}{Explicit} & \multicolumn{2}{c}{Flexible}   \\
    \cline{3-4}
       &   & Single & Multiple   \\
    \midrule
     w/o CoT &  51.4\% & 48.6\%  &  44.0\%    \\
     \systemname &  54.2\%  & 54.2\%  &  46.7\% \\
    \bottomrule
    \end{tabular}
    \vspace{-0.2cm}
    \caption{
    Success rate (\%) in real-world experiments.
    }
    \vspace{-0.3cm}
  \label{tab:real_world}
\end{table}

\section{Conclusion and Discussion}
\label{sec:conclu}

This work presents a novel 6-DoF grasp detection framework that bridges the gap between flexible instructions and robotic grasping. 
By integrating a physical-property-oriented CoT mechanism with QA-guided hierarchical reasoning, our method aligns target parsing, physical property analysis, and grasp action selection with grasp detection. The proposed multimodal LLM architecture, augmented by 3D-aware visual-textual token fusion, achieves state-of-the-art performance on a newly established benchmark \datasetname for diverse verbal-command grasping. Real-world validations further underscore its practical applicability.

While our method demonstrates robust performance under flexible instructions, some limitations persist. For instance, grasp predictions for slender objects (e.g., chopsticks) occasionally deviate from mass centers despite leveraging shape descriptors, suggesting the need to model latent physical properties beyond geometric cues. Additionally, future directions include extending the framework to sequential video inputs for dynamic scenes, such as grasping moving targets, which would require video-based grasping benchmarks to advance real-world adaptability.

\section*{Acknowledgments}
This work was supported by the National Natural Science Foundation of China (No. 62332016) and the Key Research Program of Frontier Sciences, CAS (No. ZDBS-LY-JSC001).

{
    \small
    \bibliographystyle{ieeenat_fullname}
    \bibliography{main}
}


\end{document}